\documentclass[conference]{IEEEtran}  

\IEEEoverridecommandlockouts          

\usepackage{cite}
\usepackage{amsmath,amssymb,amsfonts}
\usepackage[e]{esvect}
\usepackage{graphicx}
\usepackage{bbm}
\usepackage{siunitx} 

\usepackage{caption}

\usepackage{subcaption}
\usepackage{textcomp}
\usepackage{color}
\usepackage[draft]{hyperref}  
\usepackage[noabbrev]{cleveref}
\usepackage{multirow}
\usepackage{tabularray}
\usepackage[T1]{fontenc}
\usepackage{algorithm} 
\usepackage[noend]{algpseudocode}
\usepackage{fancyhdr}
\usepackage{dsfont}
\usepackage{censor}

\usepackage{url}
\usepackage{multicol}
\usepackage{multirow}
\usepackage{booktabs}

\usepackage{stfloats}
\usepackage{makecell}
\usepackage{easyReview}

\usepackage{xspace}

\usepackage[acronym]{glossaries}
\newacronym{rl}{RL}{Reinforcement Learning}
\newacronym{mri}{MRI}{magnetic resonance imaging}
\newacronym{ct}{CT}{Computed Tomography}
\newacronym{us}{US}{ultrasound}
\newacronym{nir}{NIR}{near-infrared}
\newacronym{icg}{ICG}{indocyanine green}
\newacronym{sbr}{SBR}{signal-to-background ratio}
\newacronym{snr}{SNR}{signal-to-noise ratio}
\newacronym{rapn}{RAPN}{robot-assisted partial nephrectomy}
\newacronym{dvrk}{dVRK}{da Vinci Research Kit}
\newacronym{pam}{PAM}{polyacrylamide}
\newacronym{pva}{PVA}{polyvinyl alcohol}
\newacronym{pla}{PLA}{polylactic acid}
\newacronym{dsc}{DSC}{DICE similarity coefficient}
\newacronym{mis}{MIS}{minimally-invasive surgery}
\newacronym{our_pipeline}{OA-NBV}{Occlusion-Aware Next-Best-View Planning for Human-Centered Active Perception on Mobile Robots}
\newacronym{nbv}{NBV}{Next-Best-View}
\newacronym{hpe}{HPE}{Human Pose Estimation}
\newacronym{nerf}{NeRF}{Neural Radiance Field}
\newacronym{prednbv}{Pred-NBV}{Pred-NBV}
\newacronym{volnbv}{Volumetric-NBV}{Volumetric-NBV}
\newacronym{hmr}{HMR}{Human Mesh Recovery}
\newacronym{leapnbv}{LEAP-NBV}{Lightweight Edge Active-Perception for Foundation-Model Next-Best-View Planning on Mobile Robots}

\usepackage[table]{xcolor}
\definecolor{pastelblue}{RGB}{173,216,230}
\definecolor{pastelpink}{RGB}{255,182,193}
\definecolor{pastelyellow}{RGB}{255,245,170}
\definecolor{pastelpurple}{RGB}{216,191,216}
\definecolor{headergray}{RGB}{235,235,235}
\definecolor{initial_camera_view}{RGB}{249,203,223}
\definecolor{best_camera_view}{RGB}{204,231,207}
\definecolor{spare}{RGB}{245,235,217}
\usepackage{array}
\usepackage{multirow}

\usepackage{graphicx}
\usepackage{array}
\usepackage[table]{xcolor}
\usepackage{adjustbox}

\usepackage{titlesec}
\titlespacing*{\section}{0pt}{*0.5}{*0.5}
\titlespacing*{\subsection}{0pt}{*0.4}{*0.4}

\title{LEAP-NBV: Lightweight Edge Active-Perception for Foundation-Model Next-Best-View Planning}

\author{
Boxun Hu$^{a}$, Jiawei Ge$^{b}$, Axel Krieger$^{b}$, Peng Wang$^{c}$, and Tinoosh Mohsenin$^{a}$
\\[3pt]
$^{a}$Department of Electrical and Computer Engineering, Johns Hopkins University, Baltimore, MD 21218, USA\\
$^{b}$Department of Mechanical Engineering, Johns Hopkins University, Baltimore, MD 21218, USA\\
$^{c}$US Army Research Laboratory, Aberdeen Proving Ground, MD 21005, USA\\[2pt]
E-mail: \{bhu29, jge9, axel, Tinoosh\}@jhu.edu; peng.wang2.civ@mail.mil
}

\begin{document}
\maketitle
\begin{abstract}

Foundation models are endowing autonomous systems with greater intelligence, enabling a more comprehensive understanding of the environment through visual perception. A representative example is \gls{hmr}, which provides useful estimates of a target's 3D pose and shape that can benefit tactical missions. However, the size and power demands of such models make them difficult to run on edge platforms and limit their real-time performance, undermining the requirements of tactical edge deployment — especially for active perception, where a mobile robot must plan its next-best view on-board and cannot offload computation under contested communications.
We present \textit{LEAP-NBV}, a lightweight active-perception framework that runs foundation-model-driven \gls{nbv} planning on-board an edge device. To this end, we distill a family of large \gls{hmr} teachers, each into a compact 32M student, with an offline mesh objective, then quantize the vision encoder to FP16 and characterize its on-device accuracy and latency. Within an occlusion-aware active-perception loop, we evaluate all configurations on the same held-out benchmark and deploy the end-to-end pipeline on an NVIDIA Jetson Xavier NX, reporting measured on-device latency and energy.
Distillation recovers 6--7\,mm of Procrustes-aligned mean per-vertex position error (PA-MPVPE) over the undistilled student on the test set. Selecting the edge-optimal compression model brings the \gls{hmr} engine to ${\sim}12$\,ms at a small accuracy cost and runs the full closed loop at 3.6\,FPS and 2.6\,J per frame, achieving a $2.0\times$ speedup and $3.0\times$ lower energy than the uncompressed model while nearly matching downstream task quality.
\end{abstract}
\section{Introduction} \label{sec:intro}

\begin{figure*}[htbp]
  \centering
  \includegraphics[width=\textwidth]{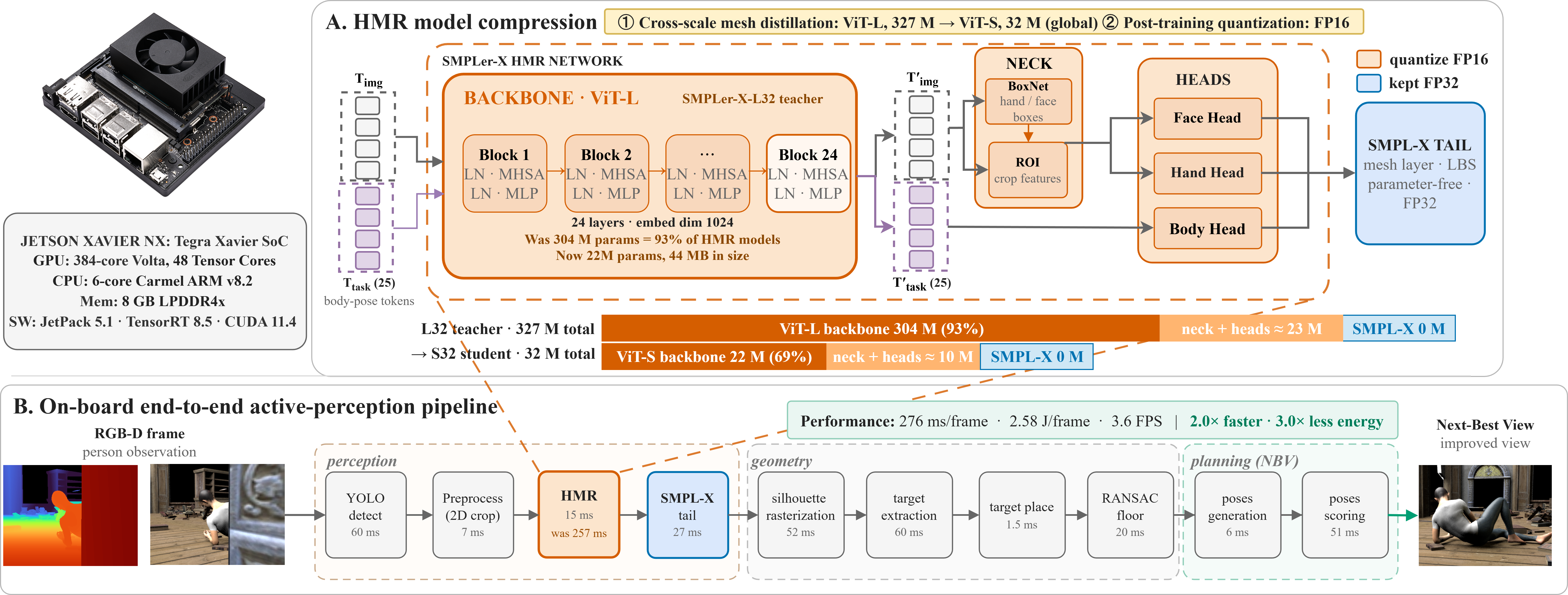}
  \caption{\textbf{\gls{leapnbv} overview.} \textbf{(A) Compression.} The ViT backbone dominates the SMPLer-X \gls{hmr} network ($304$\,M, $93\%$); \gls{leapnbv} distills it into a ViT-S student ($327{\rightarrow}32$\,M) and quantizes the backbone and heads to FP16, keeping the SMPL-X tail in FP32. \textbf{(B) On-board pipeline.} The compressed model drives the active-perception loop (detection, HMR, geometry, \gls{nbv} planning) on a Jetson Xavier NX at $276$\,ms/frame and $2.58$\,J/frame, $2.0\times$ faster and $3.0\times$ lower energy than the FP32 model.}\vspace{-16pt}
  \label{fig:system_overview}
\end{figure*}

Foundation models are growing ever larger~\cite{bommasani2021opportunities}, granting autonomous platforms increasingly capable perception, from human mesh recovery~\cite{pavlakos2019expressive} and depth estimation~\cite{yang2024depth} to broad spatial understanding~\cite{oquab2024dinov2}. Such perception, however, is most useful when the platform can act on it on its own: in tactical settings, a system must decide autonomously under degraded conditions to enable further information acquisition and timely action. A central capability is planning where to look next: a mobile robot rarely obtains a single informative view, and under occlusion it must actively plan its \gls{nbv}~\cite{kiciroglu2020activemocap} to gather more information about a target. These conditions impose a clear requirement: the robot must plan its next viewpoint autonomously and run the driving perception on-board, without an operator in the loop or reliance on cloud connectivity that is often unavailable in the field.

Humans are a primary target of interest in tactical scenarios, and \gls{hmr} provides a comprehensive estimate of a target's 3D pose and shape that is valuable for automated perception. However, mesh accuracy scales with model size: state-of-the-art \gls{hmr} backbones reach hundreds of millions to billions of parameters~\cite{cai2023smpler, yin2025smplest}, demanding compute and memory that increase inference latency. Edge platforms are tightly constrained in computation and power, so these models struggle to meet real-time requirements on-board~\cite{zhou2019edge}. The active \gls{nbv} loop compounds the problem, as it runs iteratively and issues many inferences per episode, so any per-view latency accumulates and quickly breaks real-time operation~\cite{chen2024gennbv}. Consequently, the most accurate perception models are precisely the ones that cannot sustain the on-board loop this requirement demands.

Model compression, chiefly knowledge distillation~\cite{hinton2015distilling} and quantization~\cite{jacob2018quantization}, is mature and could close the gap of \gls{hmr}, but each used alone is limited: distillation shrinks the parameter count but leaves numerical precision untouched, while quantization lowers precision but its benefit is not predictable from a device's specification, and it can also damage accuracy. Combining distillation and quantization is therefore the natural way to compress along both axes, yet their compound effect on structured 3D \gls{hmr}, under an edge budget and within the closed active-perception loop, remains poorly characterized.

To this end, we present \gls{leapnbv}, a lightweight active-perception framework for on-board \gls{hmr} foundation-model \gls{nbv} planning on edge devices (Fig.~\ref{fig:system_overview}). At its core, \gls{leapnbv} compresses a large \gls{hmr} backbone into a compact edge model through two complementary steps. First, we distill a family of large \gls{hmr} teachers each into a compact 32M student with an offline mesh objective. Second, we apply post-training quantization to reduce the model further. Finally, we characterize the accuracy–latency trade-offs across four teacher families and FP32/FP16 precisions on a held-out benchmark under controlled occlusion, select the most suitable configuration, and deploy the end-to-end pipeline on the NVIDIA Jetson Xavier NX.

The key contributions of this work are summarized as follows:
\begin{itemize}
  \item \textbf{An edge-compression approach for HMR foundation models.} We couple offline cross-scale mesh distillation with on-device post-training quantization. Distillation takes a $327$M-parameter model to a $32$M student while recovering $6$--$7$ mm PA-MPVPE over the same-size model trained without distillation, and post-training quantization further compresses and accelerates it, cutting HMR latency by ${\sim}240$ ms compared to the uncompressed model. 
  \item \textbf{A systematic accuracy--latency characterization on a deployment-realistic benchmark.} We evaluate four model variants across FP32/FP16 precisions on a held-out benchmark of $504$ unseen scenes and subjects under controlled occlusion. Characterizing the accuracy--latency trade-off directly on device, we select the edge-optimal configuration, a distilled FP16 student that runs its \gls{hmr} in ${\sim}12$\,ms at a ${\approx}1$\,mm accuracy cost, and expose a width-dependent FP16 overflow in the final normalization that collapses the large teachers but that distillation removes, leaving the deployed student numerically safe.
  \item \textbf{On-device efficiency and closed-loop task validation.} On the NVIDIA Jetson Xavier NX we measure the deployed pipeline's end-to-end per-frame energy and throughput: $2.58$\,J at $3.6$\,FPS ($276$\,ms/frame), $3.0\times$ lower energy and $2.0\times$ faster than the uncompressed model. On the closed-loop benchmark, we show that this compression keeps downstream task quality close to the teachers.
\end{itemize}

\section{Related Work}\label{sec:realated_work}

\subsection{Foundation Models for Human Mesh Recovery} Human mesh recovery estimates a parametric 3D body from a single image~\cite{pavlakos2019expressive, loper2023smpl}, recovering the pose and shape of a target. This compact 3D estimate is a useful prior for downstream autonomy: it grounds human-aware planning, interaction, and motion forecasting~\cite{kiciroglu2020activemocap, mavrogiannis2023core}, where knowing where a person is and how they are posed helps drive the system's next action. The accuracy of such estimates has advanced with the broader foundation-model trend~\cite{bommasani2021opportunities}: across visual perception, scaling Vision-Transformer backbones and training data has repeatedly improved performance~\cite{oquab2024dinov2, zhai2022scaling}. \gls{hmr} follows the same trend. SMPLer-X~\cite{cai2023smpler} and SMPLest-X~\cite{yin2025smplest} reach state-of-the-art accuracy with backbones from hundreds of millions to over a billion parameters, and their scaling studies explicitly show accuracy improving with larger models and more data. This scaling, however, ties accuracy to model size: the most accurate \gls{hmr} models carry the heaviest compute and memory footprints, which an on-board, real-time active-perception loop cannot afford. \gls{leapnbv} targets this regime, retaining foundation-model accuracy while meeting edge budgets.

\subsection{Model Compression for Efficient Inference}
Model compression reduces the cost of large networks along three main axes. Pruning removes redundant weights or structures~\cite{uttej-vlsi-2024}, though unstructured sparsity rarely yields wall-clock speedups on commodity accelerators without specialized kernels. Knowledge distillation transfers a large teacher's behavior to a compact student by matching logits~\cite{hinton2015distilling} or intermediate features~\cite{walczak2025bitmedvit}, and has mostly targeted discrete or task-specific outputs in detection~\cite{chen2017learning} and segmentation~\cite{liu2019structured} rather than the continuous 3D mesh regression that \gls{hmr} requires. Quantization lowers numerical precision, either through quantization-aware training~\cite{jacob2018quantization} or post-training calibration from a few samples~\cite{nagel2021white}, yet it is fragile on transformer-based \gls{hmr}: the large dynamic range of transformer activations can break low-precision execution~\cite{liu2021post}, so its effect must be characterized on the target hardware rather than assumed from a specification. Efficient inference for human pose and mesh estimation has been pursued mainly through lightweight backbone design~\cite{yu2021lite}. However, the joint distillation--quantization accuracy--latency trade-off for full \gls{hmr} on real edge accelerators within a closed loop remains underexplored. \gls{leapnbv} addresses this gap by compressing \gls{hmr} with cross-scale mesh distillation and characterizing quantization directly on device.

\section{Methods}\label{sec:Method}

We first set up the active-perception loop and the \gls{hmr} model that drives it, which dominates the loop's latency and power (Sec.~\ref{subsec:overview}). Because that model is too large for the edge, we present two compression steps that shrink it: cross-scale mesh distillation (Sec.~\ref{subsec:distill}) and post-training quantization (Sec.~\ref{subsec:quant}). Finally, we compile the compressed model into a TensorRT engine on the target device and integrate it into the full pipeline (Sec.~\ref{subsec:deploy}).

\subsection{Overview and problem setup}\label{subsec:overview}
We adapt the active-perception loop according to OA-NBV~\cite{hu2026oa}. At each step, the robot observes the target with an RGB-D camera, reconstructs a human mesh, forms a human-centered geometric representation, scores a set of candidate viewpoints, and moves to the selected next-best view before observing again (Fig.~\ref{fig:system_overview}B). The loop factors into three stages. \emph{Perception} detects the person through YOLOv8s~\cite{varghese2024yolov8}, preprocesses the crop, runs the \gls{hmr} model, and converts the predicted parameters into a mesh through the SMPL-X tail. \emph{Geometry} turns the mesh and the RGB-D frame into a human-centered point set. It first rasterizes the predicted mesh (vertices $V$, faces $F$) under the virtual camera into a silhouette mask and a mesh-depth map, $(M_{\mathrm{sil}}, D_{\mathrm{mesh}}) = \mathrm{Raster}(V, F; K)$. The person mask is the silhouette intersected with the dilated detection box, and the target point set is the back-projection of the masked, depth-clustered pixels:
  \begin{equation}
    P = \pi_K^{-1}\!\big(D \odot M\big), \qquad M = M_{\mathrm{sil}} \cap B,
    \label{eq:extract}
  \end{equation}
where $D$ is the aligned depth, $\odot$ the element-wise product, $B$ the detection box, and $\pi_K^{-1}$ back-projects a pixel to 3D with intrinsics $K$. We anchor the target by its robust centroid $c = \mathrm{median}(P)$ and fit the ground plane by RANSAC on the background depth. \emph{Planning} generates a set of candidate viewpoints $\mathcal{C}$ on a ground-plane ring of radius $r$ around and oriented toward the target, and selects the next-best view by maximizing a viewpoint score~\cite{hu2026oa}:
  \begin{equation}
  \begin{aligned}
    v^{\star} &= \arg\max_{v \in \mathcal{C}} \big[\, w_v\, S_v(v) + w_a\, S_a(v) + w_o\, S_o(v) \,\big], \\[2pt]
    S_v(v) &= \tfrac{1}{|P|}\textstyle\sum_i \mathbf{1}\!\left[\pi_v(p_i)\in\Omega\right],
    S_a(v) = \tfrac{1}{WH}\,\mathrm{Area}\!\left(\pi_v(\mathcal{B})\right), \\[2pt]
    S_o(v) &= \tfrac{1}{|P|}\textstyle\sum_i \mathbf{1}\!\left[\, z_v(p_i) \le
             \min_{s\in\mathcal{N}_\delta(p_i)} z_v(s) \,\right]
  \end{aligned}
  \label{eq:nbv}
  \end{equation}
where $\{p_i\}_{i=1}^{|P|}$ are the target points of $P$ projected by $\pi_v$ under candidate view $v$, $z_v(\cdot)$ the depth under that view, $\mathcal{N}_\delta(p_i)$ the background points projecting within $\delta$ pixels of $p_i$, and $w_v{=}0.03$, $w_a{=}0.14$, $w_o{=}0.83$ the default weights~\cite{hu2026oa}. The \emph{visibility} $S_v$ is the fraction of target points landing inside the $W{\times}H$ image $\Omega$, the \emph{apparent size} $S_a$ the projected area of the target's bounding box $\mathcal{B}$ as a fraction of the image, and the \emph{non-occlusion} $S_o$ the fraction not hidden by nearer background geometry. We treat the geometry and planning stages as a fixed downstream module from an occlusion-aware \gls{nbv} planner~\cite{hu2026oa}, and focus on the perception front end, which is the compute and energy bottleneck of the loop.

Within the loop, an \gls{hmr} model $f$ maps a person crop $I$ to a posed mesh. It factors into a vision-transformer encoder $g$, lightweight regression heads $h$, and a parameter-free SMPL-X layer $\mathcal{S}$:
\begin{equation}
V = f(I) = \mathcal{S}\big(h(g(I))\big), \qquad V \in \mathbb{R}^{N\times 3},
\end{equation}
where $h(g(I))$ predicts SMPL-X parameters (body/hand/face pose, shape, expression, and camera) and $\mathcal{S}$ skins them into $N{=}10{,}475$ vertices. Mesh accuracy scales with the size of $g$, whose parameters range from $22$\,M (ViT-S) to $632$\,M (ViT-H) and account for up to $93\%$ of $f$ (Fig.~\ref{fig:system_overview}A). Our goal is to obtain a single compact $f$ that preserves mesh accuracy while meeting on-board latency and energy budgets.

\begin{figure}[!t]
  \centering
  \includegraphics[width=\columnwidth]{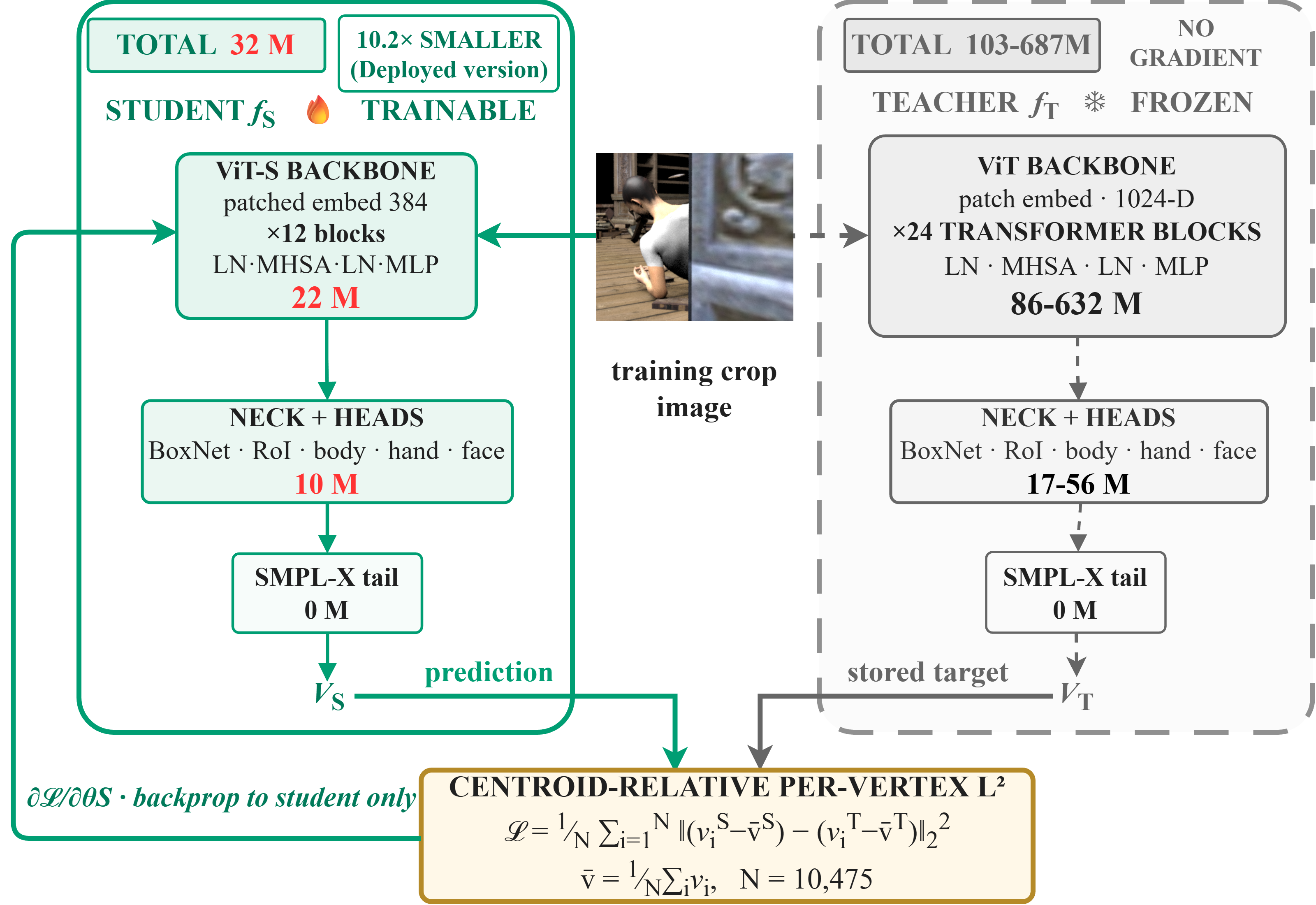}
      \caption{\textbf{Cross-scale mesh distillation.} A frozen teacher (right) is run once to precompute a target mesh $V_T$ for every training crop; the trainable student (left) predicts $V_S$ from the same crop and is optimized to match it by the centroid-relative per-vertex loss (Eq.~\ref{eq:distill}), which back-propagates to the student only. Distillation collapses each $103$--$687$\,M teacher into the same $32$\,M student ($3$--$21\times$ smaller), shrinking both the ViT backbone and the neck/heads while leaving the parameter-free SMPL-X tail unchanged.}
  \label{fig:distiallation}
\end{figure}

\subsection{Cross-scale mesh distillation}\label{subsec:distill}
We are given a family of teacher models $\{f_T\}$ of increasing size and distill each into one compact student $f_S$ (SMPLer-X-S, $32$\,M). Because the student is supervised by the teacher's mesh rather than ground-truth labels, distillation is label-free and trained on the deployment domain.
To decouple training from the large teachers, we precompute the teacher meshes offline: for every person crop $I$ in a training set $\mathcal{D}$ we store $V_T = f_T(I)$ once, and never run the teacher again during student training (Fig.~\ref{fig:distiallation}). The student is then trained to match these meshes with a translation-invariant per-vertex objective,
\begin{equation}
\mathcal{L}(I) = \frac{1}{N}\sum_{i=1}^{N}
  \big\lVert (v_i^{S}-\bar{v}^{S}) - (v_i^{T}-\bar{v}^{T}) \big\rVert_2^2,
\quad \bar{v} = \frac{1}{N}\sum_{i=1}^{N} v_i,
\label{eq:distill}
\end{equation}
where $V_S=f_S(I)$ and $\bar{v}$ is the mesh centroid. Subtracting the centroid removes the global camera translation, which otherwise dominates a raw-vertex loss; the residual measures pose and shape, exactly the quantity that the Procrustes-aligned evaluation metric rewards. One run per teacher yields one distilled student, producing a set of students we compare in Sec.~\ref{sec:exp_results}.  \vspace{-5pt}

\begin{figure}[!t]
  \centering
  \includegraphics[width=\columnwidth]{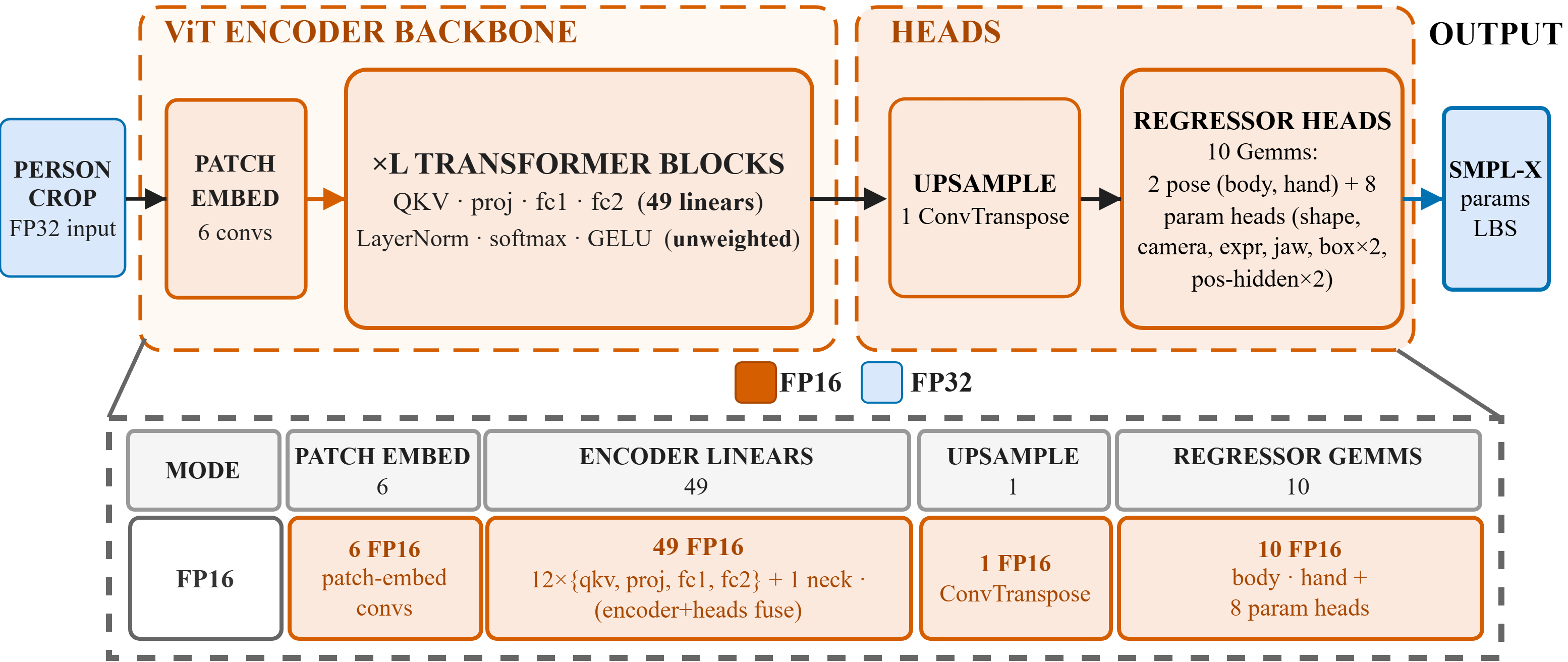}
      \caption{\textbf{On-device precision map.} Precision of the deployed FP16 engine over the $66$ weighted layers: the $6$ patch-embed convolutions, the $49$ encoder linears, the upsample, and the $10$ regressor GEMMs (body/hand pose and the eight parameter heads) all run in FP16, while the unweighted normalization, softmax, activation ops and the parameter-free SMPL-X tail are kept in FP32. This uniform half-precision assignment halves the engine's memory footprint.}
  \label{fig:quantization}
\end{figure}

\subsection{Post-training quantization}\label{subsec:quant}
Distillation shrinks the whole model; quantization then further compresses and accelerates it by lowering the numerical precision of its dominant compute. We cast every weighted layer of the compressed model to FP16 and keep the parameter-free SMPL-X layer $\mathcal{S}$ in floating point (Fig.~\ref{fig:quantization}). As a plain half-precision cast, FP16 halves the model's memory footprint and runs at real-time speed on the target Jetson Xavier NX. \vspace{-3pt}

However, half precision is not automatically safe. The final, task-token-facing normalization computes a variance over the $D$-dimensional embedding; with activations of order $\pm 14$ this sum of squares grows as ${\sim}D\cdot14^2$, which at ViT-H width ($D{=}1280$) reaches ${\approx}2.5\times10^5$ and exceeds the FP16 maximum of $65{,}504$, overflowing and collapsing the task tokens into a catastrophic mesh drift (Sec.~\ref{sec:exp_results}). The compact student sidesteps this entirely: its narrower embedding ($D{=}384$) keeps the variance well within FP16 range, so its engine is a uniform, numerically-safe half-precision cast that needs no per-layer precision surgery.\vspace{-3pt}

\subsection{On-device deployment}\label{subsec:deploy}\vspace{-3pt}
Because compiled engines are hardware- and library-specific, we build the \gls{hmr} engine directly on the target Jetson Xavier NX. The deployed model runs the encoder and regression heads as a single TensorRT engine and keeps the tail in floating point: the 6D-to-axis-angle conversion, camera-translation recovery, and the SMPL-X layer $\mathcal{S}$. This engine is dropped into the on-board pipeline of Fig.~\ref{fig:system_overview}B, where detection, preprocessing, \gls{hmr}, geometry, and \gls{nbv} scoring run back-to-back each frame. In Sec.~\ref{sec:exp_results} we characterize the resulting accuracy--latency--memory trade-offs across the model variants and their FP32/FP16 precisions, select the deployment configuration, and validate the full loop on the NVIDIA Jetson Xavier NX.
\section{Experiments and Results} \label{sec:exp_results}

\begin{figure}[!t]
  \centering
  \includegraphics[width=\columnwidth]{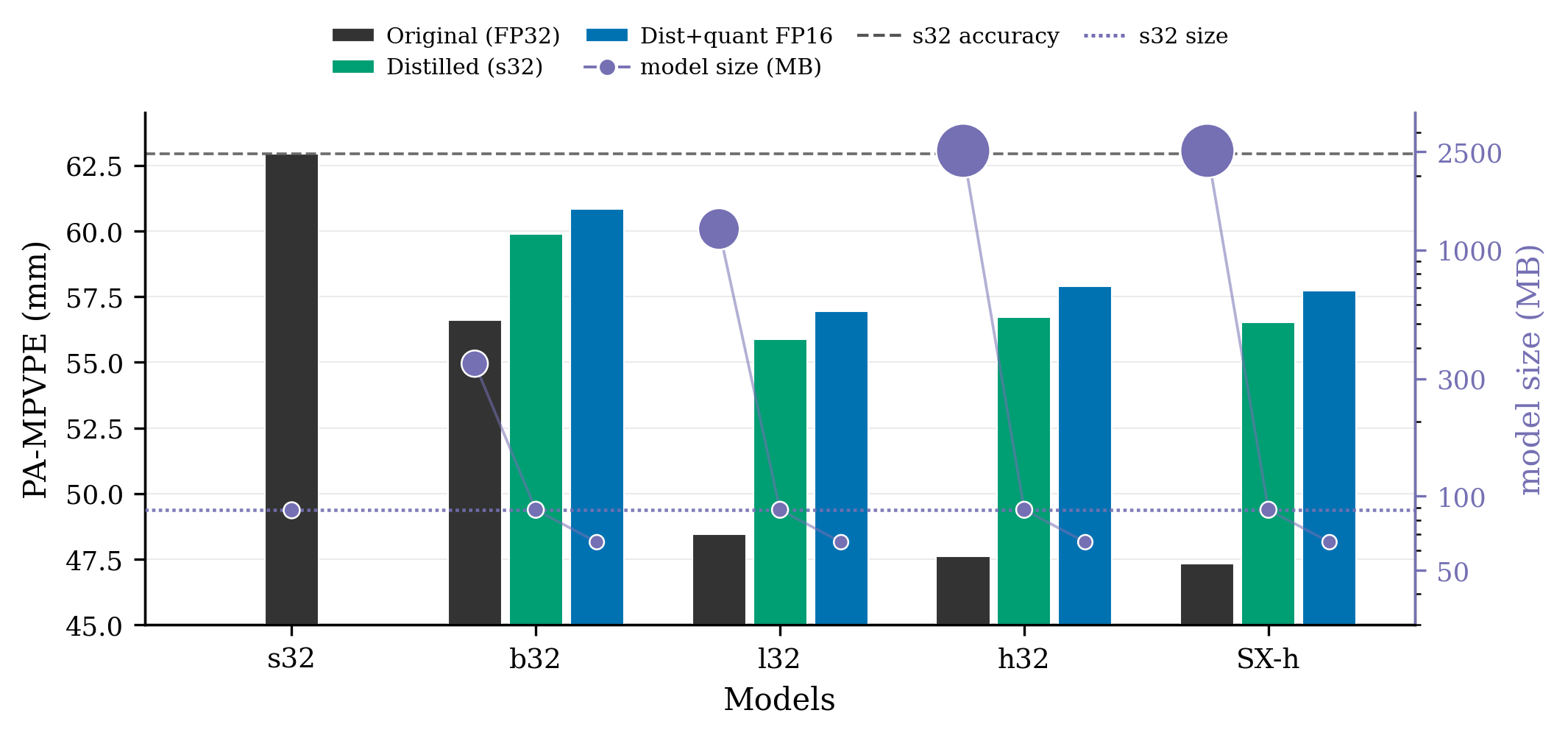}
    \caption{\textbf{Compression ladder: accuracy and model size.} For each model family, bars report single-view PA-MPVPE (left axis; lower is better) and dots report model size (right axis, log scale, with dot area also scaled to size) across three variants: the original FP32 teacher, the distilled s32 student, and that student quantized to FP16. Distillation collapses every teacher into a $32$\,M student at a small accuracy cost, and FP16 quantization shrinks it further. The dashed and dotted lines mark the baseline s32's accuracy and size; s32 has no teacher, so only its original is shown.}
  \label{fig:bars}
\end{figure}

\subsection{Experimental setup}\label{subsec:setup}

\textbf{Models.} We use the state-of-the-art SMPLer-X~\cite{cai2023smpler} checkpoints with ViT-S/B/L/H backbones (\texttt{s32}, \texttt{b32}, \texttt{l32}, \texttt{h32}; we keep the suffix from the original paper, denoting their 32-dataset training) and SMPLest-X-H (\texttt{SX-h}) as its successor~\cite{yin2025smplest}. The smallest, \texttt{s32}, is our baseline and the student architecture ($32$\,M); the four larger models serve as teachers, and \texttt{s32}$\leftarrow$X denotes the student distilled from teacher~X.

\textbf{Benchmark and metrics.} The distillation set mixes real and synthetic data: AGORA~\cite{patel2021agora} real images provide a real-image diversity anchor, while Blender-rendered scenes in the deployment domain (SMPL-X bodies placed in DISC~\cite{jeon2019disc} environments) match the target setting. The evaluation benchmark is a held-out set of these rendered scenes, with subjects and environments disjoint from training, under controlled occlusion ($20$--$60\%$ occluded). For single-view accuracy we crop each of $504$ held-out views with its ground-truth person box, isolating mesh accuracy from detection; for the closed loop we run the full active-perception pipeline over $441$ frames with the real detector (YOLOv8s~\cite{varghese2024yolov8}). Mesh accuracy is the Procrustes-aligned per-vertex error (PA-MPVPE, mm), reported single-view and in the closed loop before and after the \gls{nbv} move.  Task performance is measured detector-free from the reconstructed geometry: body coverage, how much of the target is visible, and reconstructed area, how large it appears in the image, likewise before and after the move. On device we report efficiency as \gls{hmr} and end-to-end latency (ms), throughput (FPS), board power (W) and per-frame energy (J), and model size (MB).

\textbf{Hardware.} All latency, energy, deployed-engine size, and accuracy are measured on the deployment target, an NVIDIA Jetson Xavier NX (Volta, TensorRT~8.5). The only exception is the FP16-overflow analysis of the full-scale teachers (Table~\ref{tab:overflow}), which do not deploy on the Jetson and are therefore evaluated on an NVIDIA A6000.

\begin{figure}[!t]
  \centering
  \includegraphics[width=\columnwidth]{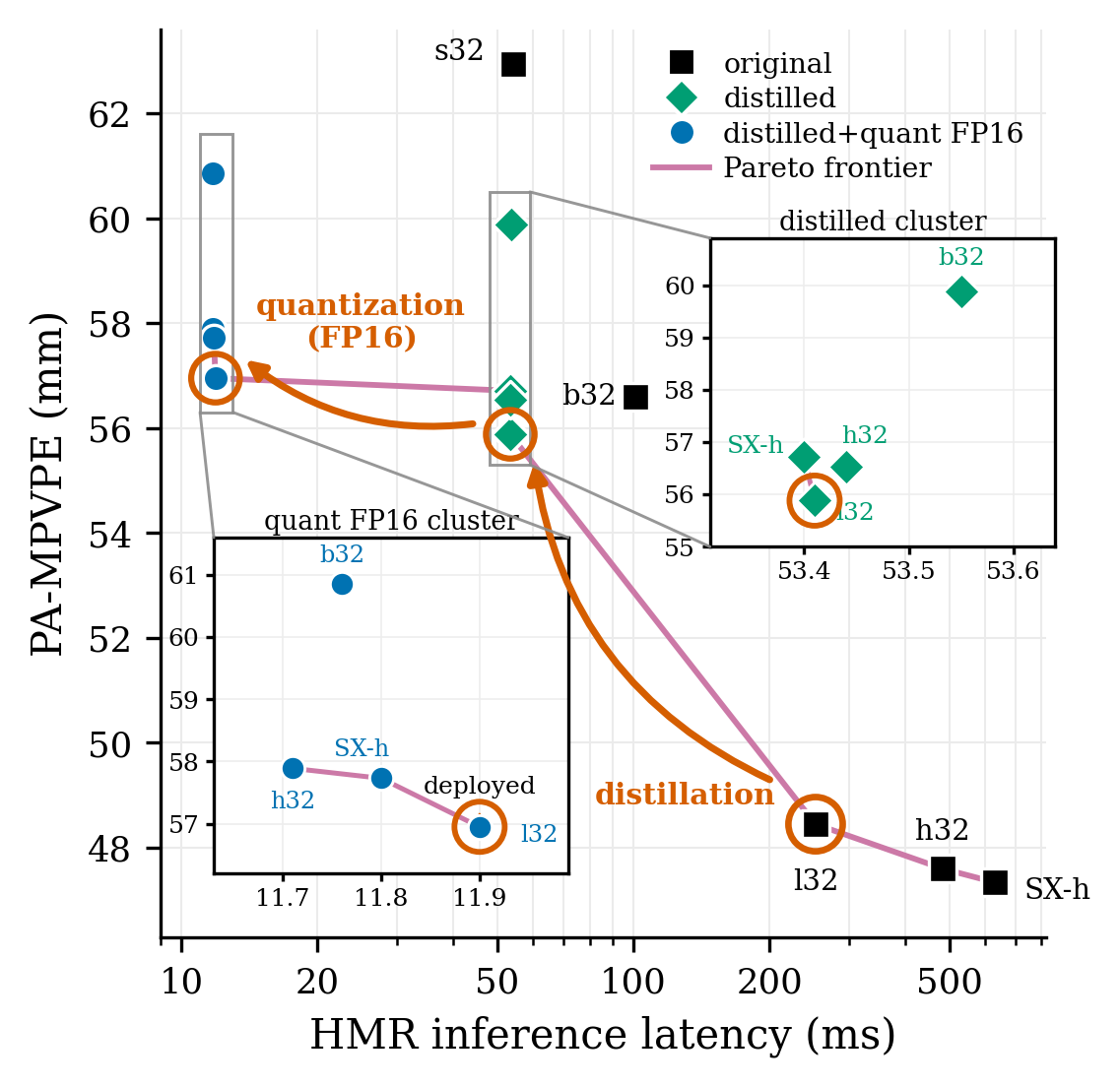}
    \caption{\textbf{Accuracy--latency Pareto on the Jetson Xavier NX.} The full compression ladder on a log latency axis: FP32 teachers (squares), distilled s32 students (diamonds), and FP16-quantized students (circles), with the Pareto frontier; the two insets zoom the distilled ($\sim$53\,ms) and quantized-FP16 ($\sim$12\,ms) clusters. Distillation and FP16 quantization move the model from the slow FP32 teachers to real-time latency at a small accuracy cost. The l32 model is ringed through both compression steps and we deploy the l32-distilled student in FP16 (circled).}\vspace{-3pt}
  \label{fig:pareto}
\end{figure}

\begin{table}[t]
\centering
\caption{\textbf{The FP16 overflow is width-dependent.} Naive FP16 casts the original model directly to half precision, uniformly across all layers. Under this cast, single-view PA-MPVPE (mm) diverges from the FP32 reference only at ViT-H width ($D{=}1280$); every model with $D{\le}1024$ matches FP32, so the deployed ViT-S student ($D{=}384$) stays well clear of the overflow.}
\label{tab:overflow}
\small
\setlength{\tabcolsep}{6pt}
\begin{tabular}{lccc}
\toprule
model & $D$ & naive FP16 & original FP32 \\
\midrule
s32  & 384  & 62.9  & 62.9 \\
b32  & 768  & 56.6  & 56.6 \\
l32  & 1024 & 48.5  & 48.5 \\
h32  & 1280 & \textbf{52.4} & 47.6 \\
SX-h & 1280 & \textbf{128.8} & 47.3 \\
\bottomrule
\end{tabular}
\end{table}

\subsection{Compression and ablation}\label{subsec:compression}
For each model family, figure~\ref{fig:bars} reports single-view PA-MPVPE (bars) and model size (dots) across three variants: the original teacher, the distilled student, and the student quantized to FP16.

\textbf{Distillation.} The original teachers confirm that mesh accuracy scales with model size: the undistilled s32 baseline reaches only $62.9$\,mm while the largest teacher (SX-h) reaches $47.3$\,mm, at $10$--$20\times$ the parameters. Distilling the larger teachers into the same $32$\,M student recovers $6$--$7$\,mm of that accuracy (e.g., L32 to $55.9$\,mm), collapsing $327$--$687$\,M models into $32$\,M while retaining most of their accuracy; the strongest students cluster near $56$\,mm. Beyond accuracy, the student is a complete SMPLer-X-S, so distillation shrinks both the backbone and the heads: from the L32 teacher, the ViT backbone drops from $304$\,M to $22$\,M and the heads from $23$\,M to $10$\,M. \vspace{-7pt}

\textbf{Quantization.} On device, the deployed FP16 engine stays within ${\sim}1$\,mm of the distilled FP32 student (Fig.~\ref{fig:pareto}). Model size falls monotonically at every step: from the FP32 teacher ($0.3$--$2.5$\,GB) to $88$\,MB (distilled FP32) and $65$\,MB (FP16), a $30$--$60\times$ reduction for the large teachers. A naive FP16 cast is not automatically safe, however: we identify a numerical overflow in the final normalization layer that appears only at ViT-H width ($D{=}1280$), collapsing SX-h to $128.8$\,mm (Table~\ref{tab:overflow}). Distillation removes it: the compact student's narrower embedding ($D{=}384$) stays within the FP16 range, so its FP16 engine is numerically safe.

\begin{figure}[!t]
  \centering
  \includegraphics[width=\columnwidth]{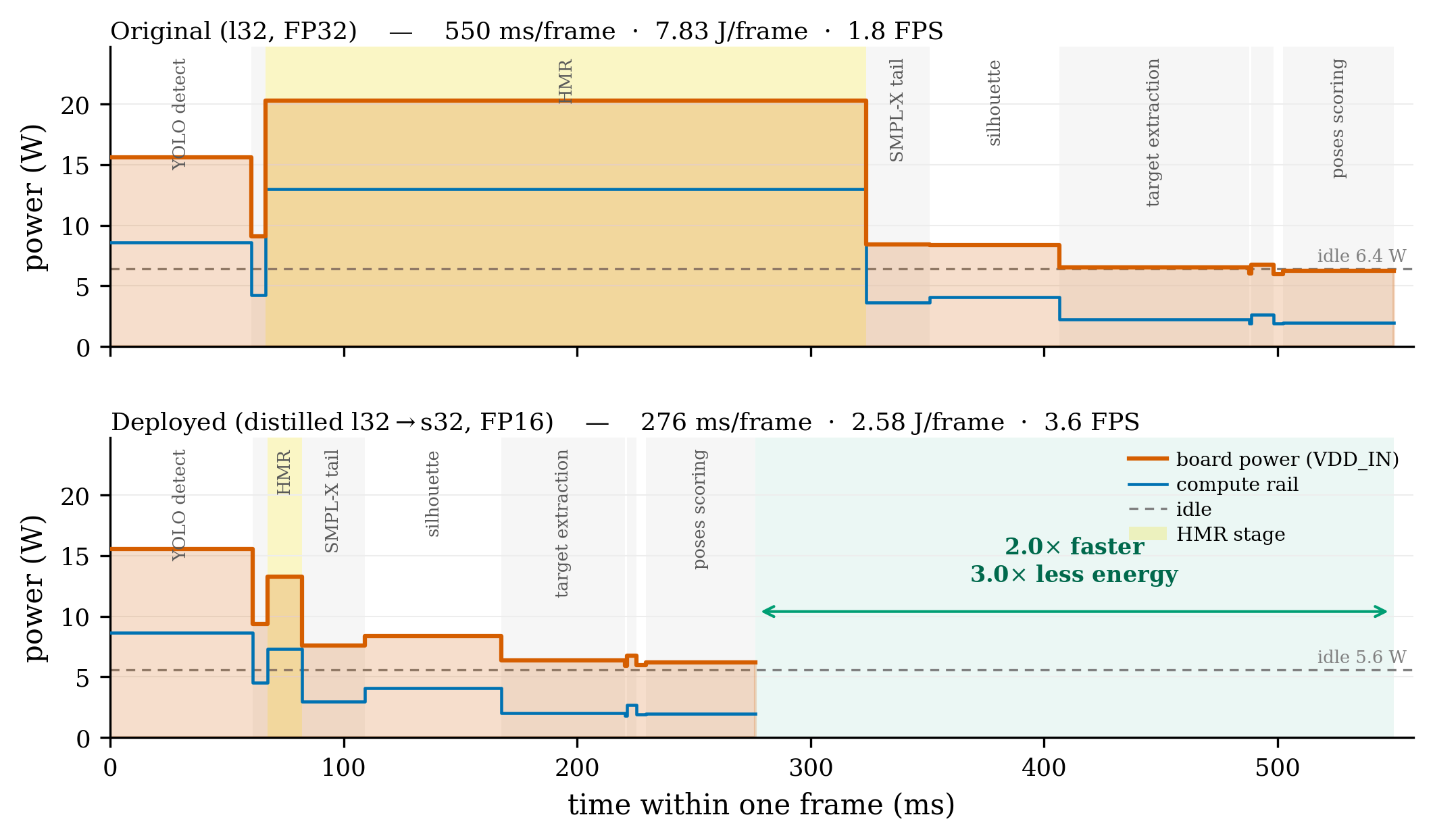}
    \caption{\textbf{End-to-end power over one frame.} Board power (VDD\_IN) and the compute rail across the stages of a single frame on the Jetson Xavier NX, for the uncompressed l32 pipeline (top) and the deployed FP16 student (bottom). The entire difference is the highlighted \gls{hmr} stage, which the deployed model runs in $12$\,ms rather than $257$\,ms; every other stage is model-independent and matches. The deployed pipeline finishes a frame in $276$,ms vs.\ $550$,ms ($2.0\times$ faster) at $2.58$,J vs.\ $7.83$,J ($3.0\times$ lower energy)}\vspace{-6pt}
  \label{fig:power}
\end{figure}

\begin{table}[b]
\centering
\caption{\textbf{Task-level performance of the \gls{nbv} loop} ($441$ frames, real detector); each cell reports the metric as \emph{initial view}\,$\to$\,\emph{after the \gls{nbv} move}. Columns: body coverage (\%), reconstructed area ($\times10^3$\,px), and closed-loop PA-MPVPE (mm); ``Deployed'' rows are distilled students in FP16. After the move, coverage and area closely match between the deployed students and their FP32 teachers, so the 2D task metrics are preserved under $30$--$60\times$ compression, while PA-MPVPE keeps a small, constant mesh-quality offset with a similar gain. The deployed configuration (\texttt{s32}$\leftarrow$\texttt{l32}) and its FP32 teacher (\texttt{l32}) are shaded for direct comparison.}
\label{tab:task}
\setlength{\tabcolsep}{4pt}
\small
\begin{tabular}{llccc}
\toprule
& model & Coverage & Area & PA-MPVPE \\
\midrule
\multirow{5}{*}{\rotatebox{90}{Original}}
 & s32 & 61.9\,$\to$\,87.9 & 11.5\,$\to$\,27.5 & 62.1\,$\to$\,55.0 \\
 & b32 & 61.9\,$\to$\,87.7 & 11.5\,$\to$\,28.1 & 54.8\,$\to$\,48.0 \\
 & \cellcolor{gray!15}l32 & \cellcolor{gray!15}61.8\,$\to$\,90.2 & \cellcolor{gray!15}11.5\,$\to$\,30.2 & \cellcolor{gray!15}49.0\,$\to$\,44.5 \\
 & h32 & 62.2\,$\to$\,88.7 & 11.4\,$\to$\,29.4 & 46.5\,$\to$\,42.4 \\
 & SX-h & 62.0\,$\to$\,88.1 & 11.5\,$\to$\,28.7 & 48.2\,$\to$\,42.6 \\
\midrule
\multirow{4}{*}{\rotatebox{90}{Deployed}}
 & s32$\leftarrow$b32 & 62.0\,$\to$\,88.3 & 11.5\,$\to$\,28.1 & 61.7\,$\to$\,54.0 \\
 & \cellcolor{gray!15}\textbf{s32$\leftarrow$l32} & \cellcolor{gray!15}\textbf{61.7\,$\to$\,88.0} & \cellcolor{gray!15}\textbf{11.4\,$\to$\,27.6} & \cellcolor{gray!15}\textbf{55.9\,$\to$\,51.0} \\
 & s32$\leftarrow$h32 & 62.1\,$\to$\,88.7 & 11.5\,$\to$\,28.8 & 56.4\,$\to$\,53.3 \\
 & s32$\leftarrow$SX-h & 62.0\,$\to$\,89.4 & 11.5\,$\to$\,28.8 & 56.4\,$\to$\,50.8 \\
\bottomrule
\end{tabular}
\end{table}

\subsection{Model and precision selection}\label{subsec:selection}
Figure~\ref{fig:pareto} places every configuration on the accuracy--latency plane measured on the Jetson. Two findings drive our choice. First, the original teachers are the most accurate but the slowest ($253$--$631$\,ms via onnxruntime) and do not build a TensorRT engine at all on the $6.8$,GB device (out of memory); they cannot be deployed as-is, so distillation is necessary. Second, among the distilled students, FP16 quantization reaches real-time latency (${\sim}12$\,ms \gls{hmr}) at a small accuracy cost, placing every deployable point at the ${\sim}12$\,ms knee of the frontier. We therefore deploy the l32-distilled student in FP16, which is the most accurate student at that latency. Notably, FP16 is numerically safe here only because distillation shrinks the embedding below the width at which the final-normalization overflow appears (Table~\ref{tab:overflow}); this underscores a broader lesson---precision behavior is model- and hardware-dependent and must be measured on the target device. \vspace{-3pt}

\subsection{End-to-end on-device deployment}\label{subsec:e2e}
Figure~\ref{fig:power} profiles the full pipeline on the Jetson Xavier NX. The deployed \gls{leapnbv} runs one frame in $276$\,ms ($3.6$\,FPS) at $2.58$\,J, sustaining the iterative move-and-re-observe loop that per-view latency would otherwise break. This is $2.0\times$ faster and uses $3.0\times$ less energy than the uncompressed l32 pipeline ($550$\,ms, $7.83$\,J, $1.8$\,FPS). The saving is concentrated in the \gls{hmr} stage, which drops from $257$\,ms at ${\sim}20$\,W to $12$\,ms; the remainder of the loop is model-independent and matches between the two. \vspace{-2pt}

We next examine how compression affects the downstream task, measured in the closed active-perception loop (Table~\ref{tab:task}). After the \gls{nbv} move, body coverage rises from $62\%$ to ${\sim}88\%$ and reconstructed area by ${\sim}2.5\times$, and closed-loop PA-MPVPE improves by a margin comparable to the FP32 teacher: the deployed student from $55.9$ to $51.0$,mm and the l32 teacher from $49.0$ to $44.5$,mm. The deployed $32$,M FP16 student reaches nearly the same downstream task outcome as the $327$--$687$,M teachers, so the $30$--$60\times$ compression that makes real-time, low-power operation possible comes at little cost to task performance. \vspace{-6pt}

\section{Conclusion}\label{sec:conclusion}
Accurate \gls{hmr} relies on foundation-model backbones that are too large and power-hungry to sustain an on-board active-perception loop. We presented \gls{leapnbv}, a lightweight framework that closes this gap by running foundation-model \gls{hmr} and next-best-view planning entirely on-device. \gls{leapnbv} distills a family of large \gls{hmr} teachers into a single $32$\,M student, recovering $6$--$7$\,mm PA-MPVPE over a same-size undistilled baseline. Characterizing post-training quantization on the target device then surfaces a key finding: a naive FP16 cast overflows the final-normalization layer and collapses the large teachers, yet the compact student's narrower embedding stays within FP16 range, making FP16 both numerically safe and the deployed choice.

Deployed end-to-end on a Jetson Xavier NX, \gls{leapnbv} runs the closed loop $2.0\times$ faster and at $3.0\times$ lower energy than the uncompressed model, while nearly matching its downstream task outcome. This shows that compression validated on-device and within the closed loop makes real-time, low-power active human perception feasible at the tactical edge. Future work will extend \gls{leapnbv} to degraded-visibility settings beyond occlusion, such as low light, and to other foundation-model perception tasks at the edge.

\let\oldthebibliography\thebibliography
\renewcommand{\thebibliography}[1]{%
  \oldthebibliography{#1}%
  \setlength{\itemsep}{0pt plus 0.2ex}%
  \setlength{\parsep}{0pt}%
}
\bibliography{bibliography, eehpc}
\bibliographystyle{ieeetr}
\end{document}